\documentclass[a4paper,twoside]{article}

\usepackage{epsfig}
\usepackage{subcaption}
\usepackage{calc}
\usepackage{amssymb}
\usepackage{amstext}
\usepackage{amsmath}
\usepackage{amsthm}
\usepackage{multicol}
\usepackage{pslatex}
\usepackage{apalike}
\usepackage{xcolor}
\usepackage{commath}
\usepackage{mathtools}
\usepackage{SCITEPRESS}     % Please add other packages that you may need BEFORE the SCITEPRESS.sty package.

\newcommand{\etal}{\textit{et al}.~}
\newcommand{\eg}{\textit{e.g}.~}

\definecolor{dkblue}{RGB}{0,64,128}
\newcommand{\best}[1]{\textcolor{dkblue}{#1}}

\newcommand{\Cityscapes}{\textsc{Cityscapes}}
\newcommand{\Pascal}{\textsc{Pascal}}

\newcommand{\eb}[1]{\scriptsize\,$\pm$\,#1\normalsize}

\begin{document}

\title{Colour augmentation for improved semi-supervised semantic segmentation}

\author{\authorname{Geoff French\sup{1}\orcidAuthor{0000-0003-2868-2237} and Michal Mackiewicz\sup{1}\orcidAuthor{0000-0002-8777-8880}}
\affiliation{\sup{1}School of Computing Sciences, University of East Anglia, Norwich, UK}
\email{\{g.french, m.mackiewicz\}@uea.ac.uk}
}

% \author{\authorname{Anonymized Author 1\sup{1}\orcidAuthor{0000-0000-0000-0000} and Anonymized Author 2\sup{1}\orcidAuthor{0000-0000-0000-0000}}
% \affiliation{\sup{1}Anonymized Institute 1}
% \email{{a1, a2}@institute1.com}
% }


\abstract{Consistency regularization describes a class of approaches that have yielded state-of-the-art results for semi-supervised classification.
While semi-supervised semantic segmentation proved to be more challenging, a number of successful approaches have been recently proposed.
Recent work explored the challenges involved in using consistency regularization for segmentation problems.
In their self-supervised work Chen~\etal found that colour augmentation prevents a classification network from using image colour statistics as a short-cut for self-supervised learning via instance discrimination.
% In their self-supervised instance discrimination work Chen~\etal found that a classification network can use image colour statistics as a short-cut, hampering the effectiveness of the network for downstream classification tasks.
% They then demonstrated that colour augmentation blocks this short-cut, forcing the network to 
Drawing inspiration from this we find that a similar problem impedes semi-supervised semantic segmentation and offer colour augmentation as a solution, improving semi-supervised semantic segmentation performance on challenging photographic imagery.}

\onecolumn \maketitle \normalsize \setcounter{footnote}{0} \vfill

\section{\uppercase{Introduction}}
\label{sec:introduction}

State-of-the-art computer vision results obtained using deep neural networks over the decade~\cite{Krizhevsky:ImageNet,He:ResNet} rely on the availability of large training sets that consist of images and corresponding annotations.
The \emph{annotation bottleneck} resulting from the manual effort involved in producing these annotations can be partially mitigated by the application of semi-supervised learning.
In contrast to traditional supervised learning in which all training samples have corresponding ground truth annotations, a semi-supervised learning algorithm is able to make used of un-annotated -- or unsupervised -- samples as well.
The presents the possibility of using a dataset in which only a subset of the training samples have corresponding annotations.

Semantic segmentation is the task of identifying the type of object or material under each pixel in an image, assigning a class to every pixel.
While efficient annotation tools~\cite{Maninis:DEXTR} can help, the cost of producing pixel-wise ground truth annotation can be significant, making the annotation bottleneck a particularly pressing issue for segmentation problems.
The progress of semi-supervised semantic segmentation has lagged behind that of semi-supervised classification.
\cite{French:SemiSupSeg} offer the challenging data distribution of semantic segmentation problems as an explanation.

The term \emph{consistency regularization}~\cite{Oliver:RealisticEval} refers to a class of approaches that have yielded state-of-the-art results for semi-supervised classification~\cite{Laine:Temporal,Tarvainen:MeanTeachers,Xie:UDA,Sohn:FixMatch} over the last few years.
\cite{French:SemiSupSeg} explore the application of consistency regularization to semantic segmentation problems, developing a successful approach based on Cutmix~\cite{Yun:Cutmix}.
They also find that plain geometric augmentation schemes used in prior semi-supervised classification approaches~\cite{Laine:Temporal,Tarvainen:MeanTeachers} frequently fail when applied to segmenting photographic imagery.

Recent work in self-supervised learning via instance discrimination trains a network for feature extraction using no ground truth labels at all.
As with consistency regularization the network is encouraged to yield similar predictions -- albeit image embeddings instead of probability vectors -- given stochastically augmented variants of an unlabelled image.
\cite{Chen:SimCLR} conducted a rigorous ablation study, finding that colour augmentation is essential to good performance.
Without out, the network in effect \emph{cheats} by using colour statistics as a short-cut for the image instance discrimination task used to train the network.
Inspired by this, we find that a similar problem can hinder semi-supervised semantic segmentation.
Our experiments demonstrate the problem by showing that it is alleviated by the use of colour augmentation.

Other recent approaches -- namely Classmix~\cite{Olsson:Classmix}, DMT~\cite{Feng:DMT} and ReCo~\cite{Liu:ReCo} -- have significantly improved on the Cutmix based results of \cite{French:SemiSupSeg}.
Our work builds on the Cutmix approach, demonstrating the effectiveness of colour augmentation.
It is not our intent to present results competitive with Classmix and DMT, thus we acknowledge that our results are not state of the art.

\section{\uppercase{Background}}
\label{sec:backg}

\subsection{Semi-supervised classification}

The key idea behind consistency regularization based semi-supervised classification is clearly illustrated in the $\pi$-model of Laine \etal~\cite{Laine:Temporal}, in which a network is trained by minimizing both supervised and unsupervised loss terms.
The supervised loss term applies traditional cross-entropy loss to supervised samples with ground truth annotations.
Unsupervised samples are stochastically augmented twice and the unsupervised loss term encourages the network to predict consistent labels under augmentation.
 
The Mean Teacher model of Tarvainen \etal~\cite{Tarvainen:MeanTeachers} uses two networks; a teacher and a student.
The weights of the teacher are an exponential moving average of those of the student. 
The student is trained using gradient descent as normal.
The teacher network is used to generate pseudo-targets for unsupervised samples that the student is trained to match under stochastic augmentation.

The UDA approach of \cite{Xie:UDA} adopted RandAugment~\cite{Cubuk:RandAugment}; a rich image augmentation scheme that chooses 2 or 3 image operations to apply from a menu of 14.
While only one network is used instead of two, we note an important similarity with Mean Teacher; just as the teacher network is used to predict a pseudo target, UDA predicts a pseudo-target for an un-augmented image that is used as a training target for the same iamge with RandAugment applied.

The FixMatch approach of \cite{Sohn:FixMatch} refines this approach further.
They separate their augmentation scheme into \emph{weak} -- consisting of simple translations and horizontal flips -- and \emph{strong} that uses RandAugment.
They predict hard pseudo-labels for \emph{weakly} augmented unsupervised samples that are used as training targets for \emph{strongly} augmented variants of the same samples.

\subsection{Semi-supervised semantic segmentation}

\cite{Hung:AdvSemiSupSeg} and \cite{Mittal:SSSHiLow} adopt GAN-based adversarial learning,
using a discriminator network that distinguishes real from predicted segmentation maps to guide learning.

\cite{Perone:SemiSupSeg} and \cite{Li:SemiSupSkin} two early applications of consistency regularisation to
semantic segmentation that we are aware of.
Both come from the medical imaging community, tackling MRI volume segmentation and skin lesion segmentation respectively.
Both approaches use standard augmentation to provide perturbation, as in the $\pi$-model~\cite{Laine:Temporal} and Mean Teacher~\cite{Tarvainen:MeanTeachers}.
\cite{Ji:IIC} developed a semi-supervised over-clustering approach that can be applied to natural photographic images, where the list of ground truth classes is highly constrained.

\cite{French:SemiSupSeg} analysed the problem of semantic segmentation, finding that it has a challenging data distribution to which the cluster assumption does not apply.
They offer this as an explanation as to why consistency regularization had not been successfully applied to semantic segmentation of photographic images.
They present an approach that drives the Mean Teacher~\cite{Tarvainen:MeanTeachers} algorithm using an augmentation scheme based on Cutmix~\cite{Yun:Cutmix}, achieving state of the art results.

% trim: [left lower right upper]
\begin{figure*}[t]
\centering
\includegraphics[width=.85\textwidth]{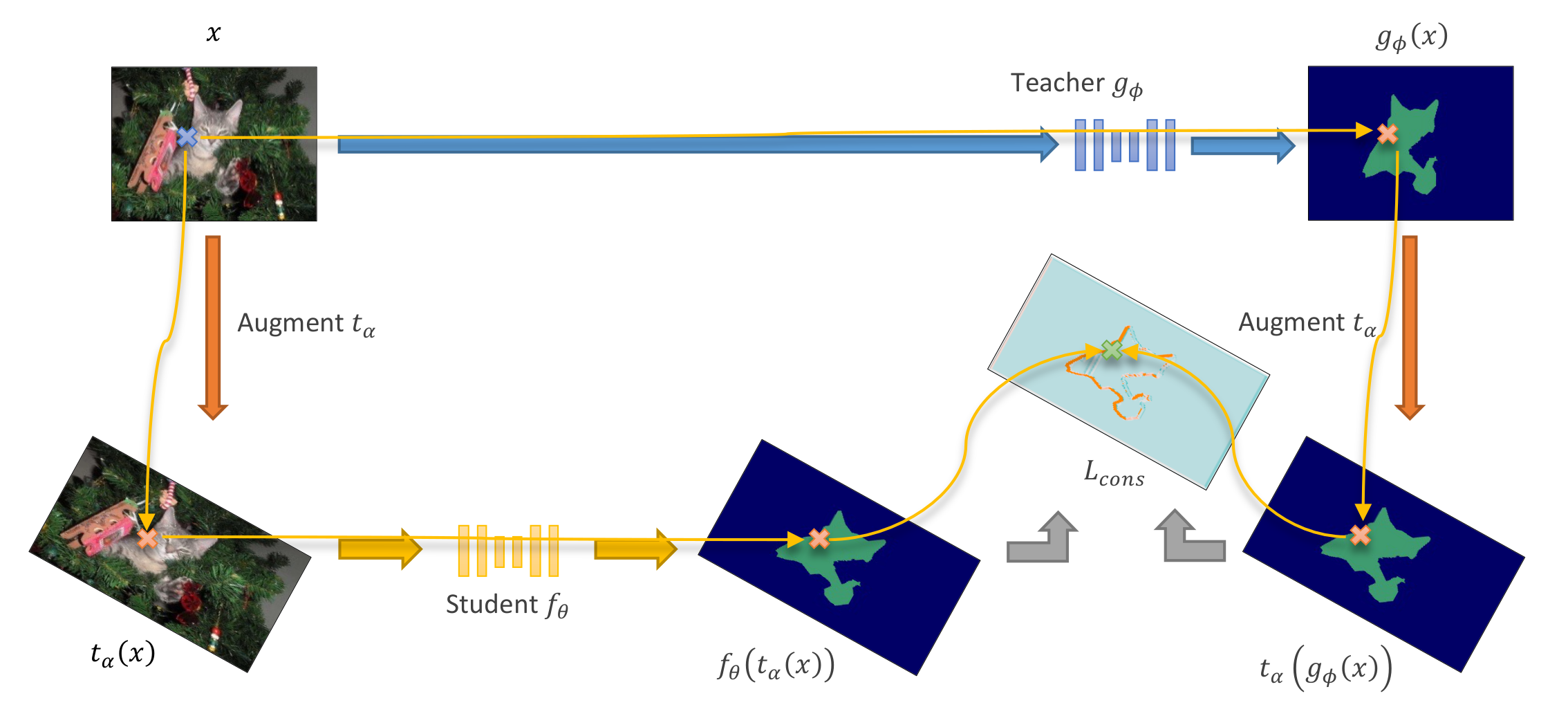}
\caption{
\label{fig:std_aug_loss}
Illustration of Mean Teacher unsupervised consistency loss driven by standard augmentation for semantic segmentation problems.
The path for a pixel on the neck of the cat leading from the input image $x$ is traced by yellows to the consistency loss map $L_{cons}$ (illustrated prior to computing the mean of the square), with the location of the pixel in each image identified by coloured crosses.
}
\end{figure*}

\subsection{Self-supervised and unsupervised learning}
\label{sec:backg:selfsup}

Approaches based on contrastive learning~\cite{Henaff:CPC,He:MoCo,Chen:MoCoV2,Chen:SimCLR} train a residual network~\cite{He:ResNet} using only unlabelled input images.
Afterwards the network backbone (consisting of convolutional layers) is frozen and a linear model is trained in a supervised fashion using it's feature representations as inputs and ground truth labels as targets.
The resulting image classifiers -- in which only the last linear layer was trained using ground truth labels -- are able to achieve ImageNet results that are competitive with those obtained by traditional supervised learning in which the whole network is trained.

In contrast to prior work~\cite{Henaff:CPC} the MoCo model of He \etal~\cite{He:MoCo} simplified contrastive learning using standard augmentation  to generate stochastically augmented variants of batches of unlabelled images.
The network is encouraged to predict embeddings that are more similar for augmented variants of the same input image than for different images.
The augmentation scheme used is very similar to the standard scheme used to train residual networks~\cite{He:ResNet} and by Mean Teacher~\cite{Tarvainen:MeanTeachers} for their ImageNet results.
Chen \etal~\cite{Chen:SimCLR} conducted a rigorous ablation study of the augmentations used for contrastive learning, assessing the effectiveness of each augmentation operation.
They found that strong colour augmentation is essential for good performance, as without it the network is able \emph{cheat} by using image colour statisticts as a short-cut to discriminate between images, rather than having to focus on image content.
Strong colour augmentation masks this signal, forcing the network to focus on the image content, extracting features suitable for accurate image classification and other downstream tasks.

We note the similarities between recent contrastive learning approaches and the Information Invariant Clustering approach of Ji \etal~\cite{Ji:IIC}, who also encourages consistency under stochastic augmentation.

The recent work of \cite{Liu:ReCo} adapt contrastive learning -- typically used for classification -- for semantic segmentation, achieving impressive results with very few labelled imgaes.

\section{\uppercase{Approach}}
\label{sec:approach}

We adopt the approach and the codebase of the semi-supervised semantic segmentation work of \cite{French:SemiSupSeg} that combines Mean Teacher~\cite{Tarvainen:MeanTeachers} with Cutmix~\cite{Yun:Cutmix}. We add colour augmentation and evaluate its' effect.
We will now describe the approaches that underpin our work.

\textbf{Semi-supervised consistency loss:} standard supervised cross entropy loss is combined with an unsupervised consistency loss term $L_{cons}$ that encourages consistent predictions under augmentation.
In a classification scenario it measures the squared difference between probability predictions from the student network $f_\theta$ and the teacher network $g_\phi$ given stochastically augmented variants $\hat{x}$ and $\tilde{x}$ of a sample $x$:

\begin{equation}
  L_{cons} = \norm{f_\theta(\hat{x}) - g_\phi(\tilde{x})}^2
\label{eqn:loss_clf}
\end{equation}
\vspace{0em}

\textbf{Semi-supervised segmentation driven with standard augmentation:} Applying standard geometric augmentation -- \eg affine transformation -- in both supervised and semi-supervised classification scenarios is straight forward.
A class-preserving transformation is drawn randomly and applied to input image use in both the supervised and unsupervised loss terms.
The predicted class probability vectors are unaffected by the transformation.

Applying geometric augmentation in segmentation scenarios requires a little more care as the classes of the pixels a semantic segmentation map -- either a ground truth segmentation map $y$ used for the supervised loss term or a segmentation map $g_\phi(x)$ predicted by a teacher network used for an unsupervised loss term -- correspond to the same pixels in the input image whose content they classify.
This input-to-target pixel correspondance must be maintained.

When training a network we must apply the augmentation $t_\alpha$ with identical parameters $\alpha$ to both the input image and segmentation map.
For our supervised loss term this means computing the loss given the networks' predictions $f_\theta(t_\alpha(x))$ given the augmented input image $t_\alpha(x)$ and the augmented ground truth $t_\alpha(y)$.
Following \cite{Perone:SemiSupSeg} this can be adapted for the unsupervised loss term in a semi-supervised scenario by applying the geometric transformation $t_\alpha$ to the input image prior to passing it to the student network and to the predicted segmentation from the teacher network (also illustrated in Figure~\ref{fig:std_aug_loss}):

\begin{equation}
  L_{cons} = \norm{f_\theta(t_\alpha(x)) - t_\alpha(g_\phi(x))}^2
\label{eqn:loss_aug}
\end{equation}
\vspace{0em}

% This correspondance must be preserved when computing consistency loss under geometric augmentation.
% A geometric transformation $t_\alpha(x)$ with parameters $\alpha$ that is applied to an input i mage $x$ -- by applying an affine transformation for example -- must also be applied -- using the same parameters -- to the segmentation map -- $t_\alpha(y)$ -- to maintain the pixel-wise correspondance between the two.
% This correspondance must be maintained when computing the consistency loss term.
% Following Perone \etal~\cite{Perone:SemiSupSeg} we randomly choose the parameters $\alpha$ for a geometric transformation $t_\alpha$ and apply it to the input image on the student side and to the predicted segmentation map on the teacher side:

% \begin{equation}
%   L_{cons} = \norm{f_\theta(t_\alpha(x)) - t_\alpha(g_\phi(x))}^2
% \label{eqn:loss_aug}
% \end{equation}
% \vspace{0em}

\textbf{Adding Cutmix to the mix:} following \cite{French:SemiSupSeg} we use Cutmix to mix two input images $x_a$ and $x_b$ to form a mixed image $x_m$ using a blending mask $m$: $x_m = x_a \odot (1 - m) + x_b \odot m$.
The same blending mask is used to mix the segmentation maps predicted by the teacher network: $y'_m = g_\phi(x_a) \odot (1 - m) + g_\phi(x_b) \odot m$ and the consistency loss term encourages the student predictions resulting from the mixed image to match the mixed segmentation maps:

\begin{equation}
  L_{cons} = \norm{f_\theta(x_m) - y'_m}^2
\label{eqn:loss_cutmix_simple}
\end{equation}

\subsection{Colour augmentation for consistency loss}
\label{sec:approach:colour_aug}

Our choice to apply colour augmentation to the unsupervised loss term in a semi-supervised semantic segmentation setting was inspired by the thorough ablation study performed by Chen \etal~\cite{Chen:SimCLR} in which they explore the effects of augmentation in the similar setting of self-supervised learning.
As stated in Section~\ref{sec:backg:selfsup}, they found that colour augmentation prevents the network from learning to use image colour statistics as a short-cut for image instance discrimination.
Colour augmentation modifies the colour statistics sufficiently to prevent the network from using them to match image instances with one another trivially, forcing the network to focus on image content.

While \cite{French:SemiSupSeg} offer the challenging data distribution present in semantic segmentation problems as an explanation as to why consistency regularization driven by standard augmentation had yielded few successes when applied to photographic image datasets such as \Pascal{} VOC 2012~\cite{Everingham:PascalVOC2012}, we offer colour statistics as an alternative explanation.

The consistency loss term in equation~\ref{eqn:loss_aug} offers the opportunity for the network to minimize $L_{cons}$ using colour statistics.
This is further illustrated in Figure~\ref{fig:std_aug_loss}, in which the yellow arrows follow a single pixel from the input image $x$ through both the student and teacher sides of the consistency loss term.
Given the care taken to maintain the input-to-target pixel correspondence as stated in Section~\ref{sec:approach}, most input pixels in $x$ (geometric augmentation can move some parts of an image outside the bounds of the image, hence correspondence for some pixels will be missing) will have corresponding pixels in the same locations in the prediction maps from both the student side $f_\theta(t_\alpha(x))$ and the teacher side $t_\alpha(g_\phi(x))$.
Given that the consistency loss term $L_{cons}$ penalises the network for giving inconsistent class predictions for each pixel, a simple way to minimize $L_{cons}$ is to predict the class of a pixel in the output segmentation maps using only the corresponding pixel in the input image, ignoring surrounding context.
Thus, the network effectively clusters the colour of individual input pixels, rather than using surrounding context to identify the type of object that the pixel lies within.

Following \cite{Chen:SimCLR} we propose using colour augmentation to prevent the network from utilizing this short-cut.
We acknowledge that \cite{Ji:IIC} applied colour augmentation in an unsupervised semantic segmentation setting.
While their codebase uses the same colour augmentation approach as \cite{He:MoCo} and \cite{Chen:SimCLR} they describe it simply as `photometric augmentation' in their paper, giving little hint that it is in fact key to the success of consistency regularization based techniques in this problem domain.

\newcommand{\RR}[1]{\raisebox{-0.25mm}{#1}}
\begin{table*}[t]
\begin{center}%
% \renewcommand{\arraystretch}{1.150000}%
% \resizebox{\linewidth}{!}{%
\setlength{\tabcolsep}{2mm}%
\begin{tabular}{@{ }llllll@{ }}
\hline
                            & \RR{\bf $\sim$1/30}     & \RR{\bf 1/8}          & \RR{\bf 1/4}          & \RR{\bf All}          \\
\RR{\# labelled}            & \RR{\bf (100)}          & \RR{\bf (372)}        & \RR{\bf (744)}        & \RR{\bf (2975)}       \\
\hline
\hline

&\multicolumn{4}{l}{\footnotesize{Results from other recent work, ImageNet pre-trained DeepLab v2 network}}    \\
%\hline
% PASTE DATA HERE
Baseline                    & ---                     & 56.2\%                & 60.2\%                & 66.0\%                \\ 
Adversarial                 & ---                     & 57.1\%                & 60.5\%                & 66.2\%                \\ 
s4GAN                       & ---                     & 59.3\%                & 61.9\%                & 65.8\%                \\ 
DMT                         & \best{\bf54.80\%}       & \best{\bf63.06\%}     & ---                   & \best{\bf68.16}\%            \\
Classmix                    & 54.07\%                 & 61.35\%               & 63.63\%               & ---                   \\
% PASTE DATA HERE
\hline

&\multicolumn{4}{l}{\footnotesize{Results from \cite{French:SemiSupSeg} and our results, ImageNet pre-trained DeepLab v2 network}}\\
%\hline
% PASTE DATA HERE
Baseline                    & 44.41\%\eb{1.11}        & 55.25\%\eb{0.66}      & 60.57\%\eb{1.13}      & 67.53\%\eb{0.35}      \\ 
Cutout                      & 47.21\%\eb{1.74}        & 57.72\%\eb{0.83}      & 61.96\%\eb{0.99}      & 67.47\%\eb{0.68}      \\ 
\quad + colour aug. (ours)  & 48.28\%\eb{1.98}        & 58.30\%\eb{0.73}      & 62.59\%\eb{0.60}      & 67.93\%\eb{0.36}      \\ 
CutMix                      & 51.20\%\eb{2.29}        & 60.34\%\eb{1.24}      & 63.87\%\eb{0.71}      & 67.68\%\eb{0.37}      \\
\quad + colour aug. (ours)  & \bf51.98\%\eb{2.77}     & \bf61.08\%\eb{0.71}   & \bf64.61\%\eb{0.57}   & \bf68.11\%\eb{0.55}   \\ 
% PASTE DATA HERE
\hline

\end{tabular}%
% }%
% \vspace*{-1mm}%
\caption[Performance on Cityscapes]{Performance (mIoU) on \Cityscapes{} validation set, presented as mean $\pm$ std-dev computed from 5 runs.
Other work: the results for 'Adversarial'~\cite{Hung:AdvSemiSupSeg} and 's4GAN'~\cite{Mittal:SSSHiLow} are taken from \cite{Mittal:SSSHiLow}.
The results for DMT~\cite{Feng:DMT} and Classmix~\cite{Olsson:Classmix} are from their respective works.
Bold results in blue colour indicate results from other works that beat our best results. Our best results are in bold.
%Per-class results can be found in the supplementary material.
}
\label{tab:results:cityscapes}
\end{center}
\end{table*}

\begin{table*}[ht!]
\begin{center}%
% \renewcommand{\arraystretch}{1.150000}%
% \resizebox{\linewidth}{!}{%
\setlength{\tabcolsep}{3mm}%
\begin{tabular}{@{ }llllll@{ }}
\hline
                       & \RR{\bf 1/100}        & \RR{\bf 1/50}         & \RR{\bf 1/20}         & \RR{\bf 1/8}          & \RR{\bf All }         \\ 
\RR{\# labelled}       & \RR{\bf (106)}        & \RR{\bf (212)}        & \RR{\bf (529)}        & \RR{\bf (1323)}       & \RR{\bf (10582)}       \\ 
\hline
\hline
&\multicolumn{5}{l}{\footnotesize{Results from other work with ImageNet pretrained DeepLab v2}}                  \\
%\hline
%Baseline              & --                    & --                    & --                    & --                    & 66.0\%                & 68.3\%                & 69.8\%                & 73.6\%               \\ 
%Semi-sup.             & --                    & --                    & --                    & --                    & 69.5\%                & 72.1\%                & 73.8\%                & --                   \\ 
%Improve               & --                    & --                    & --                    & --                    & 3.5\%                 & 3.8\%                 & 4.0 \%                & --                   \\ 

% PASTE DATA HERE
Baseline                        & --           & 48.3\%                & 56.8\%               & 62.0\%                 & 70.7\%                \\ 
Adversarial                     & --           & 49.2\%                & 59.1\%               & 64.3\%                 & 71.4\%                \\ 
s4GAN+MLMT                      & --           & 60.4\%                & 62.9\%               & 67.3\%                 & 73.2\%                \\ 
DMT                             & \best{\bf63.04\%} & \best{\bf67.15\%} & \best{\bf69.92\%}   & \best{\bf72.70\%}      & \best{\bf74.75\%}     \\ 
Classmix                        & 54.18\%      & 66.15\%               & 67.77\%              & 72.00\%                & ---                   \\ 
% PASTE DATA HERE
\hline
&\multicolumn{5}{l}{\footnotesize{Results from \cite{French:SemiSupSeg} + ours, ImageNet pre-trained DeepLab v2 network}}\\
%\hline
% PASTE DATA HERE
Baseline                        & 33.09\%      & 43.15\%               & 52.05\%              & 60.56\%                & 72.59\%               \\ 
Std. aug.                       & 32.40\%    & 42.81\%             & 53.37\%              & 60.66\%                & 72.24\%               \\ 
\quad + colour aug. (ours)      & 46.42\%    & 49.97\%             & 57.17\%              & 65.88\%                & 73.21\%               \\ 
VAT                             & 38.81\%      & 48.55\%               & 58.50\%              & 62.93\%                & 72.18\%               \\ 
\quad + colour aug. (ours)      & 40.05\%      & 49.52\%               & 57.60\%              & 63.05\%                & 72.29\%               \\ 
ICT                             & 35.82\%      & 46.28\%               & 53.17\%              & 59.63\%                & 71.50\%               \\ 
\quad + colour aug. (ours)      & 49.14\%      & 57.52\%               & 64.06\%              & 66.68\%                & 72.91\%               \\ 
Cutout                          & 48.73\%      & 58.26\%               & 64.37\%              & 66.79\%                & 72.03\%               \\ 
\quad + colour aug. (ours)      & 52.43\%      & 60.15\%               & 65.78\%              & 67.71\%                & 73.20\%               \\ 
CutMix                          & \bf53.79\%   & 64.81\%               & 66.48\%              & 67.60\%                & 72.54\%               \\ 
\quad + colour aug. (ours)      & 53.19\%      & \bf65.19\%            & \bf67.65\%           & \bf69.08\%             & \bf73.29\%            \\ 
% PASTE DATA HERE

\hline
\hline
&\multicolumn{5}{l}{\footnotesize{\cite{French:SemiSupSeg} + ours, ImageNet pre-trained DeepLab v3+ network}}\\
% PASTE DATA HERE
Baseline                        & 37.95\%      & 48.35\%               & 59.19\%              & 66.58\%                & 76.70\%               \\ 
CutMix                          & 59.52\%      & \bf67.05\%            & 69.57\%              & 72.45\%                & 76.73\%               \\ 
\quad + colour aug. (ours)      & \bf60.02\%   & 66.84\%               & \bf71.62\%           & \bf72.96\%             & \bf77.67\%            \\ 
% PASTE DATA HERE

\hline
\hline
&\multicolumn{5}{l}{\footnotesize{\cite{French:SemiSupSeg} + ours, ImageNet pre-trained DenseNet-161 based Dense U-net}}\\
% PASTE DATA HERE
Baseline                        & 29.22\%      & 39.92\%               & 50.31\%              & 60.65\%                & 72.30\%               \\ 
CutMix                          & \bf54.19\%   & \bf63.81\%            & \bf66.57\%           & 66.78\%                & 72.02\%               \\ 
\quad + colour aug. (ours)      & 53.04\%      & 62.67\%               & 63.91\%              & \bf67.63\%             & \bf74.16\%            \\ 
% PASTE DATA HERE

\hline
\hline
&\multicolumn{5}{l}{\footnotesize{\cite{French:SemiSupSeg} + ours, ImageNet pre-trained ResNet-101 based PSPNet}}\\
% PASTE DATA HERE
Baseline                        & 36.69\%      & 46.96\%               & 59.02\%              & 66.67\%                & 77.59\%               \\ 
CutMix                          & \bf67.20\%   & 68.80\%               & 73.33\%              & 74.11\%                & 77.42\%               \\ 
\quad + colour aug. (ours)      & 66.83\%      & \bf72.30\%            & \bf74.64\%           & \bf75.40\%             & \bf78.67\%            \\ 
% PASTE DATA HERE

\hline
\end{tabular}%
% }%
% \vspace*{-1mm}%
\caption[Performance on Pascal VOC 2012]{Performance (mIoU) on augmented \Pascal{}~VOC validation set, using same splits as Mittal \etal \cite{Mittal:SSSHiLow}.
Other work: the results for 'Adversarial'~\cite{Hung:AdvSemiSupSeg} and 's4GAN'~\cite{Mittal:SSSHiLow} are taken from \cite{Mittal:SSSHiLow}.
The results for DMT~\cite{Feng:DMT} and Classmix~\cite{Olsson:Classmix} are from their respective works.
Bold results in blue colour indicate results from other works that beat our best results. Our best results are in bold.
}
\label{tab:results:pascalaug}
\end{center}
\end{table*}

\begin{table*}[ht!]
\begin{center}%
% \renewcommand{\arraystretch}{1.150000}%
% \resizebox{\linewidth}{!}{%
\setlength{\tabcolsep}{3mm}%
\begin{tabular}{@{ }lllllll@{ }}
\hline
\RR{Baseline}         & \RR{Std. aug.}      & \RR{VAT}      & \RR{ICT}      & \RR{Cutout}   & \RR{CutMix}         & \RR{Fully sup.}           \\ 
\RR{(50)}         &             &         &         &         &             & \RR{(2000)}         \\
\hline
\hline
\multicolumn{6}{l}{\footnotesize{Results from \cite{Li:SemiSupSkin} with ImageNet pre-trained DenseUNet-161}}                           \\
% PASTE DATA HERE
72.85\%         & \best{\bf75.31\%} & --            & --                & --                & --                    & \best{\bf79.60\%}         \\ 
% PASTE DATA HERE
\hline
\multicolumn{6}{l}{\footnotesize{Our results: Same ImageNet pre-trained DenseUNet-161}}\\
%\hline
% PASTE DATA HERE
67.64\%         & 71.40\%         & 69.09\%     & 65.45\%     & 68.76\%     & \bf74.57\%      & 78.61\%   \\
\eb{1.83}       & \eb{2.34}         & \eb{1.38}     & \eb{3.50}     & \eb{4.30}     & \bf\eb{1.03}      & \eb{0.36}       \\
% PASTE DATA HERE

\multicolumn{6}{l}{\footnotesize{\quad + colour augmentation}}\\
%\hline
% PASTE DATA HERE
            & 73.61\%         & 61.94\%     & 50.93\%     & 73.70\%     & 74.51\%       &         \\
            & \eb{2.40}         & \eb{6.72}     & \eb{7.16}     & \eb{2.59}     & \eb{1.95}       &             \\
% PASTE DATA HERE

\hline

\hline
\end{tabular}%
% }%
% \vspace*{-1mm}%
\caption[Performance on ISIC 2017 skin lesion segmentation dataset]{Performance on ISIC 2017 skin lesion segmentation validation set, measured using the Jaccard index (IoU for lesion class). Presented as mean $\pm$ std-dev computed from 5 runs. All baseline and semi-supervised results
use 50 supervised samples. The fully supervised result ('Full') uses all 2000.
}
\label{tab:results:isic}
\end{center}
\end{table*}

\section{\uppercase{Experiments}}
\label{sec:experiments}

Our experiments follow the same procedure as \cite{French:SemiSupSeg}, using the same network architectures.
We used the same hyper-parameters, with the exception of the consistency loss weight that we will discuss in Section~\ref{sec:experiments:pascal:consweight}.

The loss term $L = L_{sup} + \gamma L_{cons}$ that we minimize is a weighted sum of a supervised cross-entropy term $L_{sup}$ with the consistency loss $L_{cons}$, with $\gamma$ as the consistency loss weight.

\subsection{Implementation}
\label{sec:experiments:implementation}

Our implementation is directly based on that of \cite{French:SemiSupSeg}.
We add colour augmentation to their implementation of standard augmentation, ICT~\cite{Verma:ICT}, VAT~\cite{Miyato:VATSemiSup}, Cutout~\cite{Devries:Cutout} and Cutmix based regularizers.
This allows us to assess its effect on a variety of regularizers across three datasets; \Cityscapes{}, \Pascal{} VOC 2012 and the ISIC Skin Lesion segmentation dataset~\cite{Codella:ISIC2017}.
Our colour augmentation randomly modifies the brightness, contrast, saturation and hue of an input image and is implemented using the \texttt{ColorJitter} transformation from the \texttt{torchvision}~\cite{PyTorch} package.

\subsection{Cityscapes}
\label{sec:experiments:cityscapes}

\Cityscapes{} is a photograpic image dataset of urban scenery captured from the perspective of a car.
Its' training set consists of 2975 images.

% Results
Our \Cityscapes{} results are presented in Table~\ref{tab:results:cityscapes} as mean intersection-over-union (mIoU) percentages, where higher is better.
The addition of colour augmentation results in a slight improvement to the CutOut and CutMix results across the board.

\subsection{Augmented Pascal VOC 2012}
\label{sec:experiments:pascal}

\Pascal{} VOC 2012\cite{Everingham:PascalVOC2012} is a photographic image dataset consisting of various indoor and outdoor scenes.
It consists of only 1464 training images, and thus we follow the lead of \cite{Hung:AdvSemiSupSeg} and augment it using \textsc{Semantic Boundaries}\cite{Hariharan:SemanticContours}, resulting in 10582 training images.

Our \Pascal{} VOC 2012 experiments evaluate regularizers based on standard augmentation, ICT~\cite{Verma:ICT} and VAT~\cite{Miyato:VATSemiSup}, Cutout and Cutmix as in \cite{French:SemiSupSeg}.

Our results are presented in Table~\ref{tab:results:pascalaug}.

\subsubsection{Consistency loss weight}
\label{sec:experiments:pascal:consweight}

We note that the effects of colour augmentation resulted in different optimal values for $\gamma$ (consistency loss weight) than were used by \cite{French:SemiSupSeg}.
When using standard geometric augmentation they found that a value of 0.003 was optimal, yielding a very slight improvement over the supervised
baseline.
Increasing $\gamma$ caused performance to drop below that of the supervised baseline.
We note that at 0.003, the consistency loss term would have little effect on training at all.
When using colour augmentation, we were able to use a value of 1 for $\gamma$; the same as that used for the more successful Cutout and CutMix regularizers.
This strongly suggests that without colour augmentation, a low value must be used for $\gamma$ to suppress the effect of the \emph{pixel colour clustering} short-cut hypothesized in Section~\ref{sec:approach:colour_aug}.

We were also able to use a value of 1 -- instead of 0.01 -- for the ICT~\cite{Verma:ICT} based regularizer when using colour augmentation.
For VAT we continue to use a weight of 0.1; we attribute this lower loss weight to the use of KL-divergence in VAT rather than mean squared error for the consistency loss.

Being able to use a single value for the consistency loss weight for all regularizers simplifies the use of our approach in practical applications.

\subsection{ISIC 2017 skin lesion segmentation}
\label{sec:experiments:isic}

The ISIC skin lesion segmentation dataset~\cite{Codella:ISIC2017} consists of dermoscopy images focused on lesions set against skin.
It has 2000 images in its training set and is a two-class (skin and lesion) segmentation problem, featuring far less variation than \Cityscapes{} and \Pascal{}.
Our results are presented in Table~\ref{tab:results:isic}.

While colour augmentation improved the performance of all regularizers on the \Pascal{} dataset when using the DeepLab v2 architecture, the results for ISIC 2017 are less clear cut.
It harms the performance of VAT and ICT, although we note that we increased the consistency loss weight of ICT to match the value used for \Pascal{}.
It yields a noticeable improvement when using standard augmentation and Cutout.
Colour augmentation increases the variance of the accuracy when using CutMix, making it slightly less reliable.
We hypothesized the the hue jittering component of the colour augmentation may harm performance in this benchmark as colour is a useful queue in lesion segmentation, so we tried disabling it when using ICT and VAT.
This did not however improve colour augmentation results.

\subsection{Comparison with other work}
\label{sec:experiments:other_work}

While we have demonstrated that colour augmentation can improve semi-supervised segmentation performance when using a simple consistency regularization based approach, we acknowledge that our results do not match those of the recent Classmix~\cite{Olsson:Classmix}, DMT~\cite{Feng:DMT} and ReCo~\cite{Liu:ReCo} approaches that use more recent semi-supervised regularizers.

We also note that \cite{Liu:ReCo} focused on situations in which a very small number of labelled samples were used.
As their work did not feature experiments with a comparable number of labelled samples to our own, we were unable to directly compare their results with ours in Tables~\ref{tab:results:cityscapes} and ~\ref{tab:results:pascalaug}.

\section{\uppercase{Discussion and conclusions}}
\label{sec:conclusions}

As observed by \cite{French:SemiSupSeg} prior work in the field of semi-supervised image classification attributed the success of consistency regularization based approaches to the \emph{smoothness assumption} \cite{Luo:SNTG} or \emph{cluster assumption} \cite{Chapelle:ClusterAssum,Sajjadi:Mutual,Shu:DIRTT,Verma:ICT}.
Their analysis of the data distribution of semantic segmentation showed that the cluster assumption does not apply.
Their successful application of an adapted CutMix regularizer to semi-supervised semantic segmentation demonstrated that the cluster assumption is in fact not a pre-requisite for successful semi-supervised learning.
In view of this, they offered the explanation that the variety of augmentation used need to provide perturbations to samples that are sufficiently varied in order to constrain the orientation of the decision boundary in the absence of the low density regions required by the cluster assumption.
CutMix succeeds due to offering more variety than standard geometric augmentation.

Our results indicate a more nuanced explanation.
The positive results obtained from adding colour augmentation to standard geometric augmentation, combined with being able to use a consistent value of 1 for the consistency loss weight for all regularizers shows that it is in fact the \emph{pixel colour clustering} short-cut that was hampering the effectiveness of standard geometric augmentation by itself, rather than a lack of variation.
Our results showing CutMix comfortably out-performing standard geometric augmentation with colour augmentation does however show that CutMix adds variety  that enables more effective learning.

The story presented by the ISIC 2017 results is less positive however.
The augmentation used to drive the consistency loss term in a semi-supervised learning scenario must be class preserving.
Modifying an unsupervised sample such that its class changes will cause the consistency loss term to encourage consistent predictions across the decision boundary, harming the performance of the classifier (see the toy 2D examples in \cite{French:SemiSupSeg} for a more thorough exploration of this).
In light of this, practitioners should carefully consider whether colour augmentation could alter the ground truth class of a sample.
We offer this as an explanation of the inconsistent effect of colour augmentation on the ISIC 2017 dataset in which the colour of lesions is an important signal.

\vfill
\section*{\uppercase{Acknowledgements}}

This work was in part funded under the European Union Horizon 2020 SMARTFISH project, grant agreement no. 773521.
Much of the computation required by this work was performed on the University of East Anglia HPC Cluster.
We would like to thank Jimmy Cross, Amjad Sayed and Leo Earl.

\bibliographystyle{apalike}
{\small
\bibliography{col_aug_sem_seg_bib_v1}}

\begin{thebibliography}{}

\bibitem[Chapelle and Zien, 2005]{Chapelle:ClusterAssum}
Chapelle, O. and Zien, A. (2005).
\newblock Semi-supervised classification by low density separation.
\newblock In {\em AISTATS}, volume 2005, pages 57--64. Citeseer.

\bibitem[Chen et~al., 2020a]{Chen:SimCLR}
Chen, T., Kornblith, S., Norouzi, M., and Hinton, G. (2020a).
\newblock A simple framework for contrastive learning of visual
  representations.
\newblock In {\em Proceedings of the 37th International Conference on Machine
  Learning}, volume 119, pages 1597--1607. PMLR.

\bibitem[Chen et~al., 2020b]{Chen:MoCoV2}
Chen, X., Fan, H., Girshick, R., and He, K. (2020b).
\newblock Improved baselines with momentum contrastive learning.
\newblock {\em arXiv preprint arXiv:2003.04297}.

\bibitem[Chintala et~al., 2017]{PyTorch}
Chintala, S. et~al. (2017).
\newblock Pytorch.

\bibitem[Codella et~al., 2018]{Codella:ISIC2017}
Codella, N.~C., Gutman, D., Celebi, M.~E., Helba, B., Marchetti, M.~A., Dusza,
  S.~W., Kalloo, A., Liopyris, K., Mishra, N., Kittler, H., et~al. (2018).
\newblock Skin lesion analysis toward melanoma detection: A challenge at the
  2017 international symposium on biomedical imaging (isbi), hosted by the
  international skin imaging collaboration (isic).
\newblock In {\em 2018 IEEE 15th International Symposium on Biomedical Imaging
  (ISBI 2018)}, pages 168--172. IEEE.

\bibitem[Cubuk et~al., 2020]{Cubuk:RandAugment}
Cubuk, E.~D., Zoph, B., Shlens, J., and Le, Q. (2020).
\newblock Randaugment: Practical automated data augmentation with a reduced
  search space.
\newblock In {\em Advances in Neural Information Processing Systems},
  volume~33, pages 18613--18624.

\bibitem[DeVries and Taylor, 2017]{Devries:Cutout}
DeVries, T. and Taylor, G.~W. (2017).
\newblock Improved regularization of convolutional neural networks with cutout.
\newblock {\em CoRR}, abs/1708.04552.

\bibitem[Everingham et~al., 2012]{Everingham:PascalVOC2012}
Everingham, M., Van~Gool, L., Williams, C. K.~I., Winn, J., and Zisserman, A.
  (2012).
\newblock The {PASCAL} {V}isual {O}bject {C}lasses {C}hallenge 2012 {(VOC2012)}
  {R}esults.
\newblock
  http://www.pascal-network.org/challenges/VOC/voc2012/workshop/index.html.

\bibitem[Feng et~al., 2021]{Feng:DMT}
Feng, Z., Zhou, Q., Gu, Q., Tan, X., Cheng, G., Lu, X., Shi, J., and Ma, L.
  (2021).
\newblock Dmt: Dynamic mutual training for semi-supervised learning.
\newblock {\em CoRR}, abs/2004.08514.

\bibitem[French et~al., 2020]{French:SemiSupSeg}
French, G., Laine, S., Aila, T., Mackiewicz, M., and Finlayson, G. (2020).
\newblock Semi-supervised semantic segmentation needs strong, varied
  perturbations.
\newblock In {\em Proceedings of the British Machine Vision Conference (BMVC)}.
  BMVA Press.

\bibitem[Hariharan et~al., 2011]{Hariharan:SemanticContours}
Hariharan, B., Arbel{\'a}ez, P., Bourdev, L., Maji, S., and Malik, J. (2011).
\newblock Semantic contours from inverse detectors.
\newblock In {\em International Conference on Computer Vision}, pages 991--998.

\bibitem[He et~al., 2020]{He:MoCo}
He, K., Fan, H., Wu, Y., Xie, S., and Girshick, R. (2020).
\newblock Momentum contrast for unsupervised visual representation learning.
\newblock In {\em Proceedings of the IEEE/CVF Conference on Computer Vision and
  Pattern Recognition}, pages 9729--9738.

\bibitem[He et~al., 2016]{He:ResNet}
He, K., Zhang, X., Ren, S., and Sun, J. (2016).
\newblock Deep residual learning for image recognition.
\newblock In {\em Proceedings of the IEEE Conference on Computer Vision and
  Pattern Recognition}, pages 770--778.

\bibitem[Henaff, 2020]{Henaff:CPC}
Henaff, O. (2020).
\newblock Data-efficient image recognition with contrastive predictive coding.
\newblock In {\em International Conference on Machine Learning}, pages
  4182--4192. PMLR.

\bibitem[Hung et~al., 2018]{Hung:AdvSemiSupSeg}
Hung, W.-C., Tsai, Y.-H., Liou, Y.-T., Lin, Y.-Y., and Yang, M.-H. (2018).
\newblock Adversarial learning for semi-supervised semantic segmentation.
\newblock {\em CoRR}, abs/1802.07934.

\bibitem[Ji et~al., 2019]{Ji:IIC}
Ji, X., Henriques, J.~F., and Vedaldi, A. (2019).
\newblock Invariant information clustering for unsupervised image
  classification and segmentation.
\newblock In {\em Proceedings of the IEEE International Conference on Computer
  Vision}, pages 9865--9874.

\bibitem[Krizhevsky et~al., 2012]{Krizhevsky:ImageNet}
Krizhevsky, A., Sutskever, I., and Hinton, G.~E. (2012).
\newblock {ImageNet Classification with Deep Convolutional Neural Networks}.
\newblock In {\em Advances in Neural Information Processing Systems 25}, pages
  1097--1105.

\bibitem[Laine and Aila, 2017]{Laine:Temporal}
Laine, S. and Aila, T. (2017).
\newblock Temporal ensembling for semi-supervised learning.
\newblock In {\em International Conference on Learning Representations}.

\bibitem[Li et~al., 2018]{Li:SemiSupSkin}
Li, X., Yu, L., Chen, H., Fu, C.-W., and Heng, P.-A. (2018).
\newblock Semi-supervised skin lesion segmentation via transformation
  consistent self-ensembling model.
\newblock In {\em British Machine Vision Conference}.

\bibitem[Liu et~al., 2021]{Liu:ReCo}
Liu, S., Zhi, S., Johns, E., and Davison, A.~J. (2021).
\newblock Bootstrapping semantic segmentation with regional contrast.
\newblock {\em arXiv preprint arXiv:2104.04465}.

\bibitem[Luo et~al., 2018]{Luo:SNTG}
Luo, Y., Zhu, J., Li, M., Ren, Y., and Zhang, B. (2018).
\newblock Smooth neighbors on teacher graphs for semi-supervised learning.
\newblock In {\em Proceedings of the IEEE Conference on Computer Vision and
  Pattern Recognition}, pages 8896--8905.

\bibitem[Maninis et~al., 2018]{Maninis:DEXTR}
Maninis, K.-K., Caelles, S., Pont-Tuset, J., and {Van Gool}, L. (2018).
\newblock Deep extreme cut: From extreme points to object segmentation.
\newblock In {\em Computer Vision and Pattern Recognition (CVPR)}.

\bibitem[Mittal et~al., 2019]{Mittal:SSSHiLow}
Mittal, S., Tatarchenko, M., and Brox, T. (2019).
\newblock Semi-supervised semantic segmentation with high-and low-level
  consistency.
\newblock {\em IEEE Transactions on Pattern Analysis and Machine Intelligence}.

\bibitem[Miyato et~al., 2017]{Miyato:VATSemiSup}
Miyato, T., Maeda, S.-i., Koyama, M., and Ishii, S. (2017).
\newblock Virtual adversarial training: a regularization method for supervised
  and semi-supervised learning.
\newblock {\em arXiv preprint arXiv:1704.03976}.

\bibitem[Oliver et~al., 2018]{Oliver:RealisticEval}
Oliver, A., Odena, A., Raffel, C., Cubuk, E.~D., and Goodfellow, I.~J. (2018).
\newblock Realistic evaluation of semi-supervised learning algorithms.
\newblock In {\em International Conference on Learning Representations}.

\bibitem[Olsson et~al., 2021]{Olsson:Classmix}
Olsson, V., Tranheden, W., Pinto, J., and Svensson, L. (2021).
\newblock Classmix: Segmentation-based data augmentation for semi-supervised
  learning.
\newblock In {\em Proceedings of the IEEE/CVF Winter Conference on Applications
  of Computer Vision}, pages 1369--1378.

\bibitem[Perone and Cohen-Adad, 2018]{Perone:SemiSupSeg}
Perone, C.~S. and Cohen-Adad, J. (2018).
\newblock Deep semi-supervised segmentation with weight-averaged consistency
  targets.
\newblock In {\em Deep Learning in Medical Image Analysis and Multimodal
  Learning for Clinical Decision Support}, pages 12--19. Springer.

\bibitem[Sajjadi et~al., 2016]{Sajjadi:Mutual}
Sajjadi, M., Javanmardi, M., and Tasdizen, T. (2016).
\newblock Mutual exclusivity loss for semi-supervised deep learning.
\newblock In {\em 23rd IEEE International Conference on Image Processing, ICIP
  2016}. IEEE Computer Society.

\bibitem[Shu et~al., 2018]{Shu:DIRTT}
Shu, R., Bui, H., Narui, H., and Ermon, S. (2018).
\newblock A {DIRT}-t approach to unsupervised domain adaptation.
\newblock In {\em International Conference on Learning Representations}.

\bibitem[Sohn et~al., 2020]{Sohn:FixMatch}
Sohn, K., Berthelot, D., Carlini, N., Zhang, Z., Zhang, H., Raffel, C.~A.,
  Cubuk, E.~D., Kurakin, A., and Li, C.-L. (2020).
\newblock Fixmatch: Simplifying semi-supervised learning with consistency and
  confidence.
\newblock In {\em Advances in Neural Information Processing Systems},
  volume~33, pages 596--608.

\bibitem[Tarvainen and Valpola, 2017]{Tarvainen:MeanTeachers}
Tarvainen, A. and Valpola, H. (2017).
\newblock Mean teachers are better role models: Weight-averaged consistency
  targets improve semi-supervised deep learning results.
\newblock In {\em Advances in neural information processing systems}, pages
  1195--1204.

\bibitem[Verma et~al., 2019]{Verma:ICT}
Verma, V., Lamb, A., Kannala, J., Bengio, Y., and Lopez-Paz, D. (2019).
\newblock Interpolation consistency training for semi-supervised learning.
\newblock {\em CoRR}, abs/1903.03825.

\bibitem[Xie et~al., 2019]{Xie:UDA}
Xie, Q., Dai, Z., Hovy, E., Luong, M.-T., and Le, Q.~V. (2019).
\newblock Unsupervised data augmentation.
\newblock {\em arXiv preprint arXiv:1904.12848}.

\bibitem[Yun et~al., 2019]{Yun:Cutmix}
Yun, S., Han, D., Oh, S.~J., Chun, S., Choe, J., and Yoo, Y. (2019).
\newblock Cutmix: Regularization strategy to train strong classifiers with
  localizable features.
\newblock In {\em Proceedings of the IEEE International Conference on Computer
  Vision}, pages 6023--6032.

\end{thebibliography}

% \section*{\uppercase{Appendix}}

% If any, the appendix should appear directly after the
% references without numbering, and not on a new page. To do so please use the following command:
% \textit{$\backslash$section*\{APPENDIX\}}

\end{document}